\documentclass{article}
\usepackage{spconf,amsmath,graphicx,multirow}
\usepackage{graphicx}
\usepackage{multirow}
\usepackage{tabularx}
\usepackage{makecell}
\usepackage{color}
\usepackage{textcomp}

\title{Automatic Music Highlight Extraction using Convolutional Recurrent Attention Networks}
\name
{Jung-Woo Ha\textsuperscript{1}, Adrian Kim\textsuperscript{1,2}, Chanju Kim\textsuperscript{2}, Jangyeon Park\textsuperscript{2}, and Sung Kim\textsuperscript{1,3}}
\address{\textsuperscript{1}Clova AI Research and \textsuperscript{2}Clova Music, NAVER Corp., Korea\\
\textsuperscript{3}Hong Kong University of Science and Technology, China}

\begin{document}
%
\maketitle
\begin{abstract}

Music highlights are valuable contents for music services. Most methods focused on low-level signal features. We propose a method for extracting highlights using high-level features from convolutional recurrent attention networks (CRAN). CRAN utilizes convolution and recurrent layers for sequential learning with an attention mechanism. The attention allows CRAN to capture significant snippets for distinguishing between genres, thus being used as a high-level feature. CRAN was evaluated on over 32,000 popular tracks in Korea for two months. Experimental results show our method outperforms three baseline methods through quantitative and qualitative evaluations. Also, we analyze the effects of attention and sequence information on performance. 
\end{abstract}
%
%
\section{Introduction}

Identifying music highlights is an important task for online music services. For example, most music online services provide first 1 minute free previews. However, if we can identify the highlights of each music, it is much better to play the highlights as a preview for users. Users can quickly browse musics by listening highlights and select their favorites. Highlights can contribute to music recommendation~\cite{cai2007scalable, su2010music, celma2010music}. Using highlights, users efficiently confirm the discovery-based playlists containing unknown or new released tracks.  

Most of existing methods have focused on using low-level signal features including the pitch and loudness by MFCC and FFT~\cite{lu2003automated, xu2009music}. Therefore, these approaches are limited to extract snippets reflecting high-level properties of a track such as genres and themes. Although extraction by human experts guarantees the high quality results, basically it does not scale.

In this paper, we assume that high-level information such as genre contribute to extract highlights, and thus propose a new deep learning-based technique for extracting music highlights. Our approach, convolutional recurrent attention-based highlight extraction (CRAN) uses both mel-spectrogam features and high-level acoustic features generated by the attention model~\cite{xu2015show}. First, CRAN finds the highlight candidates by focusing on core regions for different genres. This is achieved by setting track genres to the output of CRAN and learning to attend the parts significant for characterizing the genres. Then, th highlights are determined by summing the energy of mel-spectrogram and the attention scores. The loss of genre classification are back propagated, and weights including the attention layer are updated in the end-to-end manner. In addition, CRAN is trained in an unsupervised way with respect to finding highlights because it does not use ground truth data of highlight regions for training.  

We evaluate CRAN on 32,000 popular tracks from December 2016 to January 2017, which are served through a Korean famous online music service, NAVER Music. The evaluation dataset consists of various genre songs including K-pop and world music. For experiments, we extract the highlighted 30 second clip per track using CRAN, and conduct qualitative evaluation with the likert scale (1 to 5) and quantitative verification using ground truth data generated by human experts.
The results show that CRAN's highlights outperform three baselines including the first 1 minute, an energy-based method, and a the attention model with no recurrent layer (CAN). CRAN also outperforms CAN and models with no attention with respect to genre classification. Furthermore, we analyze the relationships between the attention and traditional low-level signals of tracks to show the attention's role in identifying highlights.

\section{Music Data Description}
We select 32,083 popular songs with 10 genres played from December 2016 to January 2017 for two months in NAVER Music. The detailed data are summarized in Table \ref{table1}. Note that some tracks belong to more than one genre, so the summation of tracks per genre is larger than the number of the data. 
The data are separated into training, validation, and test sets. Considering a real-world service scenario, we separate the data based on the ranking of each track as shown in Table \ref{table2}. We use two ranking criteria such as the popularity and the released date. We extracted a ground-truth dataset with highlights of 300 tracks by eight human experts
for quantitative evaluation,
explained in Section 4.1. The experts marked the times when they believe that highlight parts start and stop by hearing tracks. 

\begin{table}[t]
	\small \caption{\label{table1}Constitution of tracks per genre}
	\begin{center}
		\begin{tabular}{c|c|c|c|c|c}
			\hline
			Genre & \# songs & Ratio & Genre & \# songs & Ratio \\
			\hline
			Dance & 5,634 & 14.9 & Jazz & 1,649 & 4.4\\
			Ballad & 8,224 & 21.9 & R\&B & 3,619 & 9.6 \\
			Teuroteu&315& 0.8 & Indie & 3,268 & 8.7	\\
			Hiphop & 4,373 & 11.6 & Classic & 891 & 2.3 \\
			Rock & 7,135 & 19.0 & Elec & 2,511 & 6.7 \\
			\hline
			Total & 37,619 & 100 \\		
			\hline
		\end{tabular}
	\end{center}
    \vskip -0.3in
\end{table}
\begin{table}[t]
	\small \caption{\label{table2}Data separation for experiments}
	\begin{center}
		\begin{tabular}{c|c|c|c}
			\hline
			Data & Ratio(\%) & Rank range(\%) & \# of data\\
			\hline
			Training & 80 & 20 - 100 & 25,667\\
			Val / Test & 10 / 10 & 10 - 20 / 0 - 10 & 3,208 \\
			\hline
		\end{tabular}
	\end{center}
    \vskip -0.3in
\end{table}
We convert \textit{mp3} files to mel-spectrograms, which are two-dimensional matrices of which row and column are the number of mel-frequencies and time slots.
Each mel-spectrogram is generated from a time sequence sampled from an \textit{mp3} file with a sample rate of 8372Hz using librosa \cite{mcfee2015librosa}.
The sample rate was set as two times of the frequency of C8 on equal temperament(12-TET). The number of mel-bins is 128 and the fft window size is 1024, which makes a single time slot of a mel-spectrogram to be about 61 milliseconds. The input representation $\bf {x}$ is generated as follows:
\begin{enumerate}
	\item[i.]{$PT(\bf {x}) \ge 240s$: use the first 240 seconds of $\bf {x}$}
	\item[ii.]{$PT(\bf {x}) < 240s$: fill in the missing part with the last 240-$PT(\bf {x})$ seconds of a track}
\end{enumerate}
where $PT(\bf {x})$ denotes the playing time of $\bf {x}$. Therefore, we can obtain a 128 $\times$ 4,000 matrix from each track.

\section{Attention-Based Highlight Extraction}

\subsection{Convolutional Recurrent Attention Networks}
CNN has been applied to many music pattern recognition approaches~\cite{schluter2014improved, ullrich2014boundary, choi2016convolutional}. A low-level feature such as mel-spectrogram can be abstracted into a high-level feature by repeating convolution operations, thus being used for calculating the genre probabilities in the output layer. The attention layer aims to find which regions of features learned play a significant role for distinguishing between genres. The attention results later can be used to identify track highlights.
We use 1D-convolution by defining each mel as a channel to reduce training time without losing accuracy, instead of the 2D-convolution ~\cite{choi2016convolutional}. 
\begin{figure}[t]
	\begin{center}
		\centering{\includegraphics[width=0.85\columnwidth]{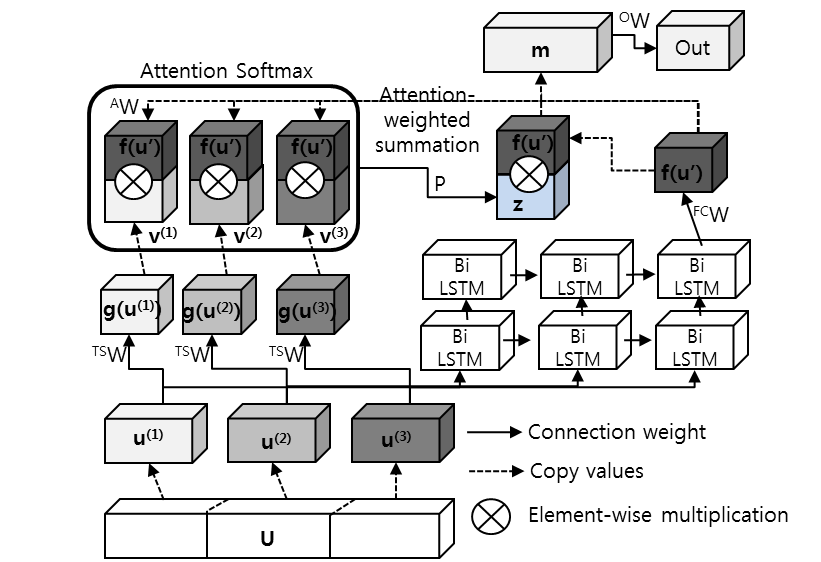}}
		\centering{\caption{\label{fig1}Structure of CRAN}} 
	\end{center}
    \vskip -0.3in
\end{figure}
Given a mel-spectrogram for a track $\bf{x}$, in specific, an intermediate feature $\mathbf{u}$ is generated through the convolution and pooling operations:
\begin{eqnarray}
\mathbf{u} = Concatenate(Maxpoolin{g^n}({Conv^k}(\mathbf{x}))) 
\end{eqnarray} 
where $n$ and $k$ denote the numbers of pooling and convolution layers. We use the exponential linear unit (elu) as a non-linear function \cite{clevert2015fast}.
After that, $\mathbf{u}$ is separated into a sequence of $T$ time slot vectors, $\mathbf{U}=\{ {\mathbf{u}^{(t)}}\} _{t = 1}^T$, which are fed into bidirectional LSTM~\cite{hochreiter1997long, graves2012long}. Then, we obtain a set of $T$ similarity vectors, $\mathbf{V}$, from the $tanh$ values of ${\mathbf{u}^{(t)}}$ and of a vector transformed from the output of LSTM, $\mathbf{u'}$:
\begin{eqnarray}
\mathbf{u'} = BiLSTM(\mathbf{U})\\
\mathbf{V} = \{ {\mathbf{v}^{(t)}}\} _{t = 1}^T= g(\mathbf{U}) \otimes f(\mathbf{u}') \label{eqn1}\\
g(\mathbf{U}) = \tanh ({}^{TS}W\mathbf{U}) \\
f(\mathbf{u}') =  Re(\tanh ({}^{FC}W\mathbf{u}'), T)
\end{eqnarray}
where $\otimes$ denotes element-wise multiplication. $Re(\mathbf{x}, T)$ is a function which makes $T$ duplicates of $\mathbf{x}$. ${}^{TS}W$ and ${}^{FC}W$ are the weight matrices of the time separated connection for attention and the fully connected layer (FC1) to the output of LSTM, as shown in Fig. \ref{fig1}. 
\begin{figure*}[h]
	\begin{center}
		\centering{\includegraphics[width=1.8\columnwidth]{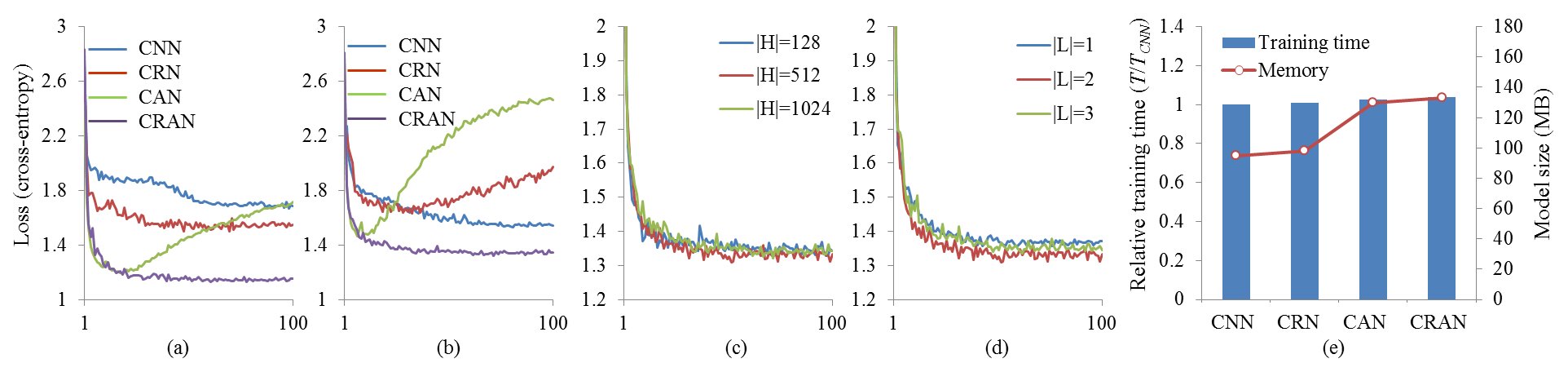}}
		\centering{\caption{\label{fig4}Model performance for various model parameters}} 
	\end{center}
    \vskip -0.3in
\end{figure*}
\begin{figure*}[h]
	\begin{center}
		\centering{\includegraphics[width=1.8\columnwidth]{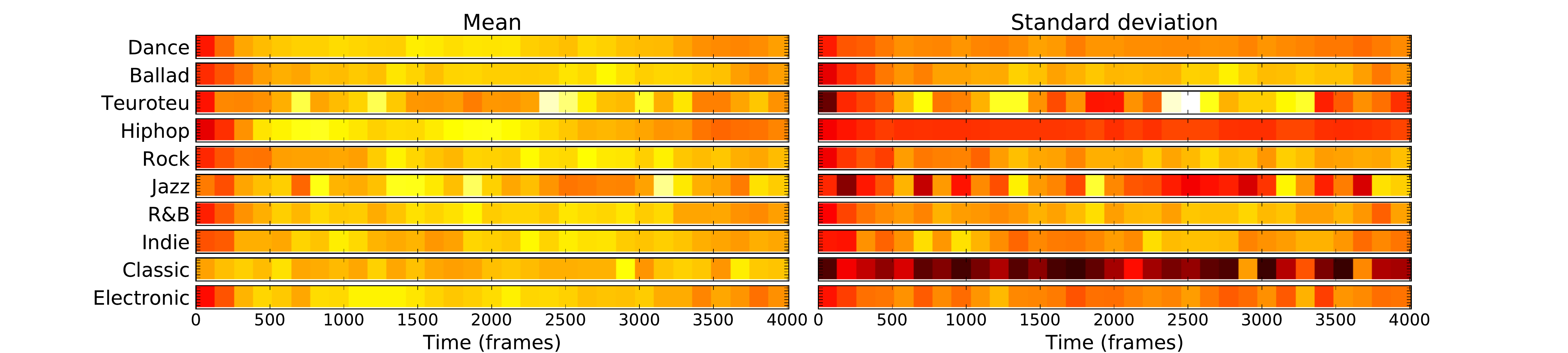}}
		\centering{\caption{\label{fig5} Distribution of attention scores according to genres. Values decrease in the order of white, yellow, red, and black.}} 
	\end{center}
    \vskip -0.3in
\end{figure*}
\begin{figure*}[h]
	\begin{center}
		\centering{\includegraphics[width=1.8\columnwidth]{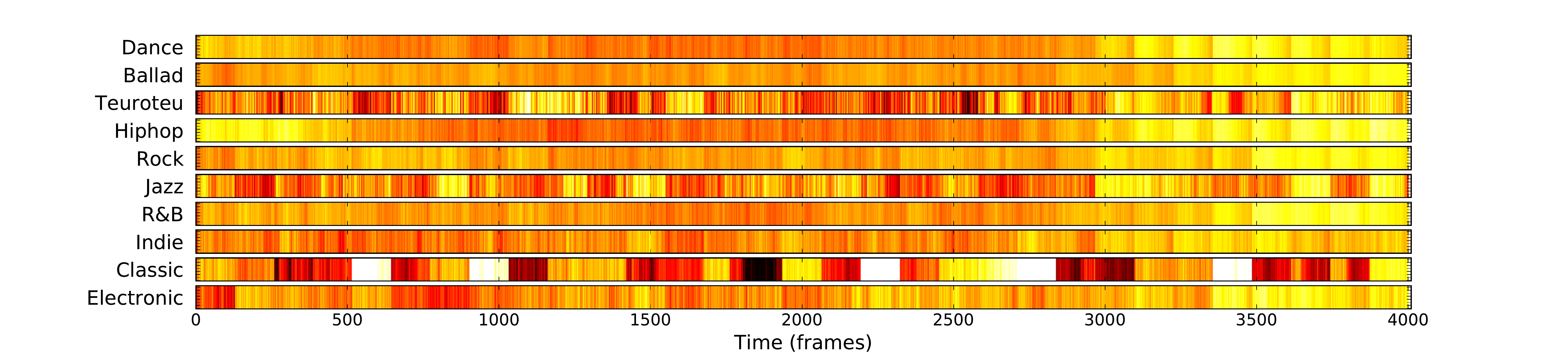}}
		\centering{\caption{\label{fig7} Correlation coefficient between attention scores and energy of mel-spectrograms for time frames per genre}} 
	\end{center}
    \vskip -0.3in
\end{figure*}
CRAN uses the soft attention approach~\cite{nam2016dual}. The attention score of $\{ {\mathbf{u}^{(t)}}\}$ is the softmax value of $\{ {\mathbf{v}^{(t)}}\}$ using a two layer-networks:
\begin{eqnarray}
{\alpha _i} = Softmax \{ \tanh ({}^AW{\mathbf{v}^{(i)}})\} 
\end{eqnarray}
where ${}^AW$ is the weight matrix of the connection between similarity vectors and each node of the attention score layers. 
Then, $\mathbf{z}$ is calculated by the attention score-weighted summation of the similarity vectors of all time slots: 
\begin{eqnarray}
\mathbf{z} = P\sum\nolimits_{t = 1}^T {{\alpha _t}{v^{(t)}}}
\end{eqnarray}
where $P$ is a matrix for dimensionality compatibility. 
Then, the context vector $\mathbf{m}$ is obtained by element-wise multiplication between the tanh values of $\mathbf{z}$ and of the FC vector:
\begin{eqnarray}
\mathbf{m} = \tanh (\mathbf{z}) \otimes \tanh ({}^{FC}W\mathbf{u'})
\end{eqnarray}
Finally, the probability of a genre $y$ is defined as the softmax function of $\mathbf{m}$. The loss function is the categorical cross-entropy~\cite{deng2006cross}. 

\subsection{Extracting Track Highlights}
We use the mel-spectrogram and the attention scores of a track together for highlight extraction. The highlight score of each frame is computed by summing the attention scores and the mean energies:
\begin{eqnarray}
{{\tilde e}^{(n)}} = \gamma {\alpha _n} + \frac{{(1 - \gamma )}}{{128}}\sum\nolimits_{i = 1}^{128} {e_i^{(n)}}  \\
{H^n} = \beta \sum\nolimits_{s = 0}^{S - 1} {{{\tilde e}^{(n + s)}}}  + (1 - \beta )\left( {\Delta {e^{(n)}} + {\Delta ^2}{e^{(n)}}} \right)
\end{eqnarray}
where $e_i^{(n)}$ and $S$ denote the energy of the $i$-th mel channel in the $n$-th time frame and the duration of a highlight. $\beta$ and $\gamma$ are arbitrary constants in (0, 1). $\Delta {e^{(n)}}$ and ${\Delta ^2}{e^{(n)}}$ denote the differences of ${e^{(n)}}$ and $\Delta {e^{(n)}}$, and they enable the model to prefer the rapid energy increment. 

\section{Experimental Results}
\subsection{Parameter Setup and Evaluation Methodology}
Hyperparameters of CRAN are summarized in Table~\ref{table3}. We compare the highlights extracted by CRAN to those generated by the method summing the energy of mel-spectrogram, the first one minute snippet (F1M), and convolutional attention model without a recurrent layer (CAN). In addition, CRAN and CAN are compared to models without attention for genre classification, called CRN and CNN, respectively.  

\begin{table}[t]
	\small \caption{\label{table3}Parameter setup of CRAN}
	\begin{center}
		\begin{tabular}{l}
			\hline
 - Convolution \& pooling layers: 2 \& 1, 4 pairs \\
 - \# and size of filters: 64 \& [3, 3, 3, 3] \\
 - Pooling method and size: max \& [2, 2, 2, 2] \\
  - \# and node size of LSTM layers: 2 \& 512 \\
 - Dropout (recurrent / fully connected): 0.2 / 0.5\\
 - Number and node size of FC layers: 2, [500, 300] \\
 - Optimizer: Adam~\cite{kingma2014adam} (LR:0.005, decay: 0.01)\\
 - $\beta$, $\gamma$: 0.5, 0.1 \\
			\hline	
		\end{tabular}
	\end{center}
    \vskip -0.3in
\end{table}

We compare the highlights extracted by CRANs to those by the three baselines. We define two metrics for the evaluation. One is the time overlapped between the ground-truth and extracted highlights. The other is the recall of extracted highlights. Given a track $\mathbf{x}$ and an extracted highlight $H$, two metrics are defined as follows:
\begin{eqnarray}
O(\mathbf{x},H) = PT(GT(\mathbf{x}) \cap H)\\
Recall(\mathbf{x},H) = \left\{ {\begin{array}{*{20}{c}}
	{1,\,if\,O(\mathbf{x},H) > \,0.5 \times PT(H)}\\
	{0,\,otherwise\,\,\,\,\,\,\,\,\,\,\,\,\,\,\,\,\,\,}
	\end{array}} \right.
\end{eqnarray}
where $PT(\mathbf{x})$ and $GT(\mathbf{x})$ denotes the playing time and the ground truth highlight of $\mathbf{x}$. In addition, five human experts rated the highlights extracted by each model in range of [1, 5] as the qualitative evaluation. 

\begin{table}[t]
	\small \caption{\label{table4}Comparison of quantitative performance}
	\begin{center}
		\begin{tabular}{c|c|c||c}
			\hline
			~ Models ~ & ~ Overlap (s) ~ & ~ Recall ~ & ~ Qual ~ \\
			\hline
			First 1 minute (F1M) & 6.96$\pm$10.70 & 0.258 & 1.793\\  
			Mel-spectrogram & 19.76$\pm$10.5 & 0.796 & 4.256\\
			CAN & 21.47$\pm$9.84& 0.857 & 4.256 \\
			CRAN & \textbf{21.63$\pm$9.78} & \textbf{0.860} & \textbf{4.268}\\
			\hline
		\end{tabular}
	\end{center}
    \vskip -0.3in    
\end{table}
\subsection{Quantitative and Qualitative Evaluations }
Table \ref{table4} presents the results. CRAN yields the best accuracy with respect to both qualitative and quantitative evaluations. This indicates that the high-level features improves the quality of the extracted music highlights. Interestingly, we can find that using F1M leads to very poor performance even if its playing time is twice. It indicates that the conventional preview is needed to be improved using the automatic highlight extraction-based method for enhancing user experience.  

Table~\ref{table5} presents the results with respect to overlap and recall according to genres. In Table~\ref{table5}, values denote the overlapped time. Overall, CRAN yields a little better performance compared to CAN and outperforms the mel energy-based method and F1M. It indicate that the attention scores are helpful for improving the highlight quality in most genres. Interestingly, all models provide relatively low performance on hiphop and indie genres, resulting from their rap-oriented or informal composition. 

\begin{table}[t]
	\small \caption{\label{table5}Comparison of mean overlapped time per genre}
	\begin{center}
		\begin{tabular}{c|c|c|c|c|c}
			\hline
			Genre & Size & F1M & Mel & CAN & CRAN \\
			\hline
			Dance & 57 & 6.56 & 20.72& 22.37 & \textbf{22.40}\\
			Ballad  &  113 & 3.41& 22.14& 23.37 & \textbf{23.75}\\
			Teuroteu & 5 & 18.8 & 20.0 & 21.20 & \textbf{22.34}\\
			Hiphop & 42 & 13.33& 14.52& 15.79 & \textbf{15.81}\\
			Rock & 27 & 7.15& 19.78& 22.11 & \textbf{22.40}\\
			Jazz & 6 & 10.0& 20.33& \textbf{20.83} & \textbf{20.83}\\
			R\&B & 53 & 7.56& 18.89& \textbf{21.77} & 21.58\\
			Indie Music & 12 & 13.25& \textbf{18.75} & 18.08 & 18.08\\
			Classical & 5 & 0.0 & 20.0 & 25.50 & \textbf{25.75}\\
			Electronic & 9 & 7.22& 17.33& 17.11 & \textbf{18.67}\\
			\hline
		\end{tabular}
	\end{center}
    \vskip -0.3in    
\end{table}

\begin{table}[t]
	\small \caption{\label{table6}Comparison of quantitative performance}
	\begin{center}
		\begin{tabular}{c|c|c|c|c}
			\hline
			~ Recall@3 ~ & ~ CNN ~ & ~ CAN ~ & ~ CRN ~ & CRAN\\
			\hline
			Popularity & 0.804 & 0.898 & 0.858 & \textbf{0.918}\\
			NewRelease & 0.802 & 0.831 & 0.791 & \textbf{0.871}\\
			\hline
		\end{tabular}
	\end{center}
\vskip -0.3in    
\end{table}

\subsection{Genre Classification and Hyperparameter Effects}
We investigate the effects of the attention mechanism on genre learning and classification performance. Recall@3 was used as a evaluation metric, considering the ambiguity and similarity between genres~\cite{panagakis2008music, silla2007automatic}. Table~\ref{table6} depicts the classification performance of each model. As shown in Table \ref{table6}, the attention mechanism considerably improves the performances as at least 0.05 on two test datasets. In addition, CRAN provides better accuracy compared to CAN, and it indicates that sequential learning is useful for classifying genres. 

Fig. \ref{fig4} shows the classification performance for model types and parameters with respect to time and accuracy. From Figs. \ref{fig4}(a) and (b), the usage of both sequential modeling and the attention mechanism prevents overfitting comparing the loss of CRAN to other models. It is interesting that the number and the hidden node size of the recurrent layers rarely contributes to improve the loss of the model from Fig. \ref{fig4} (c) and (d). The use of attention does not require much training time while the use of recurrent layers slightly increases the model size, as shown in Fig. \ref{fig4}(d).

\subsection{Attention Analysis}
Fig. \ref{fig5} presents the distribution of the mean and the variance of the attention scores derived from CRAN per genre. As shown in Fig. \ref{fig5}, time slots with a large attention score vary by genre. In particular, ballad, rock, and R\&B tracks show similar attention patterns. Hiphop and classical genres show relatively low standard deviation of attention scores due to their characteristics~\cite{gall2005music}. This result indicates the attention by CRAN learns the properties of a genre.

Fig. \ref{fig7} presents the correlation coefficient between the attention scores and the energy of mel-spectrogram for each genre. we find that the regions with higher energy in the latter of a track are likely to be a highlight. In addition, high energy regions are obtained larger attention scores through entire time frames in classical music, compared to other genres. We infer that high-level features can play a complement role for extracting information from tracks, considering the different patterns between attention scores and low-level signals. 

\section{Concluding Remarks}
We demonstrate a new music highlight extraction method using high-level acoustic information as well as low-level signal features, using convolutional recurrent attention networks (CRAN) in an unsupervised manner. We evaluated CRAN on 32,083 tracks with 10 genres. Quantitative and qualitative evaluations show that CRAN outperforms baselines. Also, the results indicate that the attention scores generated by CRAN pay an important role in extracting highlights. 
As future work, CRAN-based highlights will be applied to Clova Music service, the AI platform of NAVER and LINE.

\bibliographystyle{IEEEbib}
\bibliography{mhe}

\end{document}